%% file: acl_latex.tex
\documentclass[11pt]{article}

\usepackage[final]{acl}

\usepackage{times}
\usepackage{latexsym}

\usepackage{xcolor}

\usepackage[T1]{fontenc}

\usepackage[utf8]{inputenc}

\usepackage{microtype}

\usepackage{inconsolata}

\usepackage{graphicx}
\usepackage{multirow} 
\usepackage{comment} 
\usepackage{booktabs}
\usepackage{xfp}
\usepackage{amsmath}

\usepackage{hyperref}

\usepackage{setspace}
\usepackage[ruled,linesnumbered]{algorithm2e}
\usepackage{adjustbox}
\usepackage{soul}
\usepackage{subcaption}
\usepackage{enumitem}
\usepackage{bm}
\usepackage{amssymb}
\usepackage[labelfont=bf]{caption}  
\usepackage{booktabs} 
\usepackage{thmtools}
\usepackage[flushleft]{threeparttable}
\usepackage{pifont} 
\usepackage{tabularx}

\usepackage{arydshln}

\title{Understanding the Role of Prompt Template in Knowledge Distillation for Safety Alignment}

\author {
    Anjila~Budathoki\textsuperscript{\textrm{1}},
    Manish~Dhakal\textsuperscript{\textrm{1}},
    Benjamin~M.~Ampel\textsuperscript{\textrm{2}},
    Yi~Ding\textsuperscript{\textrm{1}}\thanks{Corresponding author} \\
    \textsuperscript{\textrm{1}}University of Tennessee, Knoxville\\
    \textsuperscript{\textrm{2}}Georgia State University \\
    {\tt\small {\{abudatho, mdhakal1\}@vols.utk.edu,  bampel@gsu.edu}, yding@utk.edu } \\
}

\author {
    Anjila~Budathoki\textsuperscript{\textrm{1}},
    Manish~Dhakal\textsuperscript{\textrm{1}},
    Benjamin~M.~Ampel\textsuperscript{\textrm{2}},
    Yi~Ding\textsuperscript{\textrm{1}} \\
    \textsuperscript{\textrm{1}}University of Tennessee, Knoxville\\
    \textsuperscript{\textrm{2}}Georgia State University \\
    {\tt\small {\{abudatho, mdhakal1\}@vols.utk.edu,  bampel@gsu.edu}, yding@utk.edu } \\
}

\begin{document}
\maketitle
\begin{abstract}
Prior research has demonstrated that the choice of prompt template during Supervised Fine-Tuning (SFT) significantly impacts the robustness of safety alignment afterwards. However, the influence of template selection during Knowledge Distillation (KD) from teacher to student remains largely unexplored.
Thus, we fill this gap by analyzing how different template configurations influence the pre-existing safety alignment of the student.
We observe a significant degradation of safety alignment present in the aligned base instruct-tuned model. 
Specifically, we find that utilizing chat templates renders the model more compliant with harmful queries compared to a non-chat template. 
These findings are consistent across three models: LLaMA, Gemma and Qwen model families and are evaluated across multiple safety benchmarks. We further show that using a non-chat template during distillation better preserves the base student's internal representations, while chat template distillation induces a larger representational shift.\footnote{Code:\url{https://github.com/anjilab/role-of-prompt-template-in-kd}}

\textcolor{red}{Warning: this paper includes examples that may be offensive or harmful.}

\end{abstract}

\input{content/Introduction}
\input{content/LiteratureReview}

\input{content/Method}

\input{content/Result}

\input{content/Analysis}
\input{content/Conclusion}

\input{content/Limitations}
\input{content/Acknowledgements}


\bibliography{acl_latex}
\clearpage
\appendix
\input{content/Appendix}

\end{document}

%% file: content/Introduction.tex
\section{Introduction}

\begin{figure*}[!t]
    \centering
    \includegraphics[width=\textwidth]{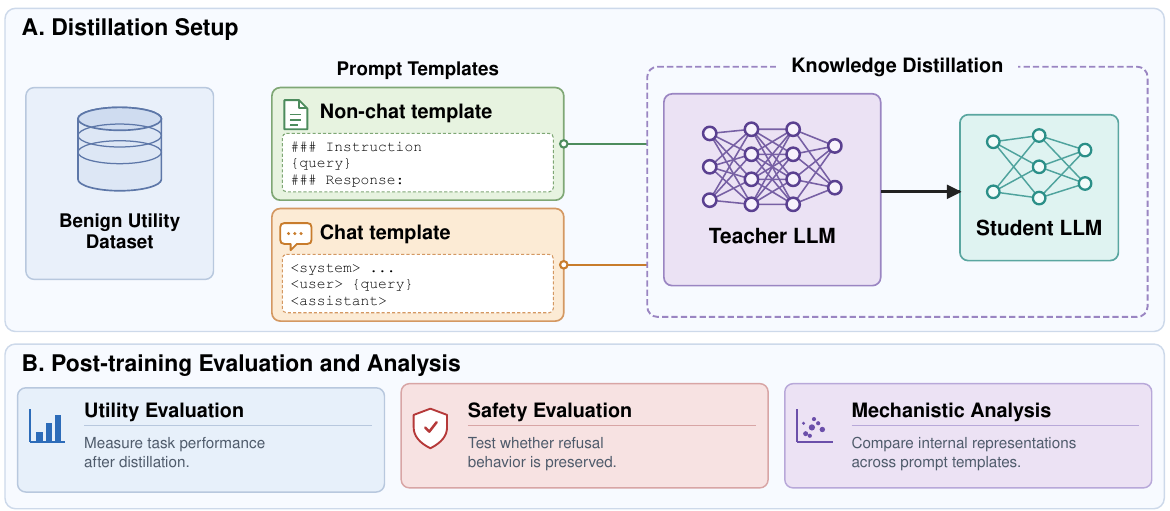}
    \caption{
\textbf{Overview of the experimental pipeline.}
We distill student models on benign instruction-following data under two formatting conditions: non-chat and chat. The pipeline compares how these training-time prompt templates affect the resulting student model during downstream safety and utility evaluation.
}
    \label{fig:kd_pipeline}
\end{figure*}

Knowledge Distillation (KD) \cite{hinton2015distilling} is an effective technique for compressing large models into smaller, efficient student models while retaining performance. In the context of Large Language Models (LLMs), KD is commonly used to obtain compact models that are easier to deploy under computational constraints. As these models are increasingly integrated into practical applications such as healthcare and autonomous systems~\citep{alabbasy2023compressing, agand2024knowledge}, considerations of robustness and safety become increasingly important. In particular, deployed distilled models are expected to exhibit safe and harmless behavior, including refusing harmful, malicious, or policy-violating requests. To encourage such behavior, modern instruction-tuned models typically undergo alignment processes that embed refusal capabilities before any downstream adaptation~\cite{askell2021general,ouyang2022training}. A key open question is how the safety alignment of student models changes during further training via KD, and which training-time factors govern its preservation or degradation.

In particular, the choice of prompt template~\cite{lyu2024keeping} during training is one such factor. Prompt templates specify how inputs are formatted for the model, often as strings with placeholders filled by user queries, instructions, and assistant responses. Modern LLMs~\cite{touvron2023llama, jiang2023mistral7b} primarily utilize \textit{chat templates}, which standardize user interactions via specific control tokens (e.g., \texttt{<|user|>}, \texttt{[INST]}); we contrast these with \textit{non-chat templates}, which present the same content without these conversational control tokens. Hereafter, we use \textit{chat templates} to refer to templates that wrap inputs with special conversational control tokens, and \textit{non-chat templates} to refer to task-style formats that omit these tokens. Prior work shows that such formatting choices can substantially affect safety behavior during SFT~\citep{lyu2024keeping,jiang2025chatbug,wang2024loss}; however, their role in KD remains empirically underexplored.

Unlike SFT, KD exposes the student to teacher-generated soft targets, an additional signal that may interact with template formatting in ways that distinctly affect the student's safety representations. This raises an important question about the interplay between distillation, template formatting, and safety retention:

\textit{How do chat and non-chat prompt templates affect the preservation of safety-aligned refusal behavior when student models are distilled on benign downstream tasks?}

In this work, we answer this question by systematically investigating how the format of prompt templates during training shapes the distilled student model's susceptibility to harmful queries~\cite{qi2024safety,arditi2024refusal}. We distill student models on benign instruction-following data under two controlled template conditions: chat and non-chat, and evaluate all models using the standard chat template at inference, ensuring that any observed safety differences reflect training-time choices alone. Beyond output-level evaluation, we conduct a mechanistic analysis to examine whether chat-template KD shifts the student’s internal refusal representations or whether the behavioral gap is a surface-level artifact. We further study \textit{prompt template mixing} by varying the proportion of chat versus non-chat samples during training to characterize how safety and utility scale with chat template exposure. The overall pipeline is shown in Figure~\ref{fig:kd_pipeline}.

Our key findings are summarized as follows:
\begin{itemize}
    \item  Knowledge distillation on benign dataset can erode the pre-existing safety alignment of aligned student models across model families.

    \item  Prompt formatting modulates the severity of this degradation: using a chat template during KD consistently leads to a higher Attack Success Rate (ASR) than non-chat KD under identical training data.
    
    \item The student-side training template, not the teacher's output distribution, is the primary driver: using the chat template only for the teacher causes minimal safety loss, whereas using it for the student produces consistent regression.

    \item The degradation is not limited to output behavior: chat-template KD shifts the student's internal refusal direction away from the base model and reduces its ability to separate harmful from harmless prompts, whereas non-chat KD largely preserves both.

    \item Safety is more sensitive to chat template exposure than utility. As the proportion of chat template samples increases, ASR generally rises, indicating amplified safety degradation, whereas utility improves only modestly.
\end{itemize}

%% file: content/LiteratureReview.tex
\section{Related work}

\paragraph{Knowledge Distillation.}
\citet{hinton2015distilling} originally demonstrated that soft targets (i.e., a full probability distribution) encode richer information than hard labels (i.e., a discrete label). This enabled the efficient transfer of knowledge from a large deep neural network to smaller ones. Recently, research has shifted towards distilling the knowledge of LLMs into more compact student models \cite{gu2024minillm, ko2024distillm, wang2025abkd}. Notably, \citet{gu2024minillm} introduced a distillation objective for generative models (adopted by many subsequent methods) which introduced reverse KL divergencefor improved KD in generative LLMs. However, despite the advancement in task-centric KD that prioritizes task utility (e.g., preserving accuracy or fluency), how these processes impact safety alignment is currently under-studied.

\paragraph{Safety Alignment in LLMs.}
To align LLMs with human values, models often adopt a two-stage post-training pipeline of SFT and alignment-tuning (e.g., Reinforcement Learning from Human Feedback, Direct Preference Optimization) 
\cite{ouyang2022training, bai2022training, rafailov2023direct}. These methods have demonstrated that careful alignment can embed the ability to reject harmful instructions (known as safety guardrails) within an LLM. Prior work has also employed KD as a mechanism to enforce safety alignment in student models \citet{yang2024distillseq}. However, literature has found that several vulnerabilities exist in the safety guardrails of open-access LLMs \citep{yi2024vulnerability}. These vulnerabilities lead to jailbreak attacks, which allow malicious users to use the LLM in unintended ways \citep{yi2024vulnerability}. It is unclear if distilling models on benign downstream tasks can further degrade these safety guardrails in previously aligned models.

\paragraph{Prompt Templates.}

Prompt templates, which determine how instructions, user inputs, and assistant responses are serialized before being passed to a model, are often treated as implementation details, but these formatting choices meaningfully shape model behavior and safety. During instruction fine-tuning, chat templates can reduce context awareness, influencing how models attend to input \citep{wang2024loss}. They also affect safety: task-style fine-tuning and safety-oriented testing better preserve safe behavior \citep{lyu2024keeping}, while certain chat template designs induce unexpected behavioral failures \citep{jiang2025chatbug}. Similar vulnerabilities also appear in VLMs, where Role-Modality Attacks exploit dialogue roles and modality placement \citep{shayegani2025misalignedrolesmisplacedimages}. Despite these findings, the role of prompt templates in knowledge distillation remains underexplored; we address this gap by systematically comparing chat and non-chat templates during benign KD and measuring their impact on both utility and safety.

%% file: content/Method.tex
\section{Methodology}

In this section, we present our approach for analyzing the impact of prompt templates on safety alignment during Knowledge Distillation (KD). Our pipeline, illustrated in Figure \ref{fig:kd_pipeline}, consists of two stages: (1) constructing instruction datasets using distinct prompt formatting strategies (chat vs. non-chat), and (2) fine-tuning a student model using a balanced KD objective.

\subsection{Prompt Formatting}
\label{sec:prompt_formatting}
To examine the role of structural cues during distillation, we design two controlled formatting conditions for the instruction dataset $\mathcal{D}$. While modern instruction-tuned models typically require specific chat templates (e.g., \texttt{<|begin\_of\_text|>}, \texttt{<|start\_header\_id|>}) to maintain state and role \citep{huggingface_transformers_chat_templates_2025}, it is unclear if these tokens aid or hinder the transfer of safety representations during KD.

We use standard definition of two template configurations, as described below (see e.g. in Table \ref{tab:prompt-template}):
\begin{itemize}
    \item \textbf{Chat Template:} We use each model family's native chat template, including all special control tokens, as the chat-format baseline for instruction following.
    \item \textbf{Non-chat Template:} We strip all model-specific control tokens, presenting the input as raw text. This isolates the semantic content of the instruction from the structural priors enforced by the template.
\end{itemize}

Importantly, regardless of the training template, all models are evaluated using the standard chat template at inference to reflect real-world deployment conditions (details in \S\ref{sec:inference-setting})

\subsection{Distillation Objective}

Given the formatted dataset $\mathcal{D} = \{(x, y)\}$, we fine-tune the student model $q_\theta$ by combining a standard supervised learning objective with Knowledge Distillation from the teacher $p$. The total loss function balances the ground-truth alignment with the transfer of the teacher's probability distribution:

\begin{equation*}
\mathcal{L} = (1 - \alpha)\mathcal{L}_{CE} + \alpha\mathcal{L}_{KD}
\label{eq:objective}
\end{equation*}

The supervised component, $\mathcal{L}_{CE}$, enforces alignment with the gold reference tokens, while $\mathcal{L}_{KD}$ minimizes the forward KL divergence between the teacher and student logits:
\begin{align*}
    \mathcal{L}_{CE} &= \mathbb{E}_{(x,y) \sim \mathcal{D}} \left[ -\log q_\theta(y | x) \right] \\
    \mathcal{L}_{KD} &= \mathbb{E}_{x \sim \mathcal{D}, y \sim p( \cdot | x) } \left[ \log p(y | x) - \log q_\theta(y | x) \right]
\end{align*}

We set $\alpha = 0.5$ to assign equal weight to task performance and knowledge transfer. Although our primary experiments use forward KL  as the standard distillation 
objective, we additionally evaluate reverse KL and combined FKL+RKL objectives 
to assess whether the observed template effects generalize across distillation 
objectives (Appendix~\ref{sec:other_distillation_objective}).

\subsection{Experimental Setup}
In this section, we detail our experimental setup for investigating how prompt template choice during distillation influences the preservation of safety alignment of the student models.


\subsubsection{Models.}
We employ three widely used open-weight model families: LLaMA-3 \cite{grattafiori2024llama}, Gemma-2 \cite{team2024gemma}, and Qwen-2.5 \cite{qwen2.5_2024}. Following standard KD protocols \cite{gu2024minillm, ko2024distillm}, we use larger Instruct variants as \textbf{teachers} ($\mathcal{M}_T$) and smaller Instruct variants as \textit{students} ($\mathcal{M}_S$): \textbf{LLaMA-3.1-8B-Instruct} $\rightarrow$ \textit{LLaMA-3.2-3B-Instruct}, \textbf{Gemma-2-9B-IT} $\rightarrow$ \textit{Gemma-2-2B-IT}, and \textbf{Qwen2.5-7B-Instruct} $\rightarrow$ \textit{Qwen2.5-3B-Instruct}. These model pairs were selected due to their rigorous pre-training filtration and post-training safety alignment, providing a strong baseline for measuring safety degradation.

\subsubsection{Datasets.}
\paragraph{Instruction Following.}
We perform distillation with the \texttt{databricks-dolly-15k} dataset \cite{gu2024minillm}, a standard benchmark for instruction following.

\paragraph{Safety Evaluation.}
To measure safety alignment, we evaluate on four adversarial benchmarks: AdvBench~\cite{zou2023universal} and JailbreakBench~\cite{chao2024jailbreakbench} for harmful-instruction attacks, and HarmBench~\cite{mazeika2024harmbench} and SORRY-Bench~\cite{xie2025sorry} for broader harmful-behavior coverage. These datasets are used for evaluation only; please refer to Appendix ~\ref{safety_eval} for a detailed description of each.

\subsubsection{Training and Evaluation}
\paragraph{Training}
We fine-tune all student models using LoRA~\cite{hu2022lora} rank $r=8$. The training is conducted on a single node equipped with 4 $\times$ NVIDIA RTX 4090 GPUs. For a comprehensive overview of hyperparameters, including learning rates and batch sizes, please refer to the Appendix \ref{sec:training_settings}.

\paragraph{Evaluation Setting.}
We assess model performance along two dimensions: task utility and safety alignment. For task utility, we report the average ROUGE-L score across the four general-domain instruction-following sets. For safety alignment, we compute Attack Success Rate (ASR) on AdvBench and JailbreakBench using the WalledEval framework~\cite{gupta2024walledeval} with \texttt{LLaMA-3-Guard-8B} as the judge, and evaluate HarmBench and SORRY-Bench using their respective provided judges.

\paragraph{Inference Setting.}
\label{sec:inference-setting} Crucially, during evaluation, we strictly adhere to the standard chat template for all models (including those trained with non-chat templates). This follows the recommended inference-time usage of modern instruction-tuned LLMs~\cite{grattafiori2024llama,team2024gemma} and ensures that our evaluation reflects the practical safety of deployed models. Therefore, any observed safety differences across conditions can be attributed to the training-time template choice, rather than to differences in the evaluation format.

%% file: content/Result.tex
\section{Results}
\input{assets/table/result_prompt_template}

\label{sec:result}

We analyze the safety behavior of student models after knowledge distillation (KD) under two training-template conditions: (1) chat template and (2) non-chat template. During evaluation, all models are tested using the standard chat template, reflecting the expected deployment setting. 

\subsection{Distillation generally erodes safety.}
We observe that standard KD on benign instruction-following data degrades the safety alignment in the student models. As shown in Table \ref{tab:result_prompt_template_safety_only}, all three distilled student models show increased ASR on most safety benchmarks relative to their corresponding base student models. In some cases, the vulnerability of the student models has almost tripled with maximal degradation (HarmBench for LLaMA and SORRY-Bench for Gemma). We further show in Appendix~\ref{sec:other_distillation_objective} that the same template-driven safety degradation persists under alternative distillation objectives.



\subsection{Template Choice Shapes Safety Degradation}

Our main finding is that the prompt template used during KD strongly affects safety 
preservation. When the student is distilled using the standard chat template, the 
resulting model shows larger increases in ASR than with the non-chat template. For 
example, on SORRY-Bench, the Gemma student shows a \textbf{+35.32\%} ASR increase 
relative to its base student model. Similarly, on HarmBench, we observe a 
\textbf{+27.57\%} increase for LLaMA and a \textbf{+13.12\%} increase for Qwen. 
These results suggest that chat-template KD is associated with greater erosion of 
pre-existing refusal behavior.

While prior work has shown that SFT can degrade safety alignment~\cite{lyu2024keeping, 
jiang2025chatbug}, we show that similar template sensitivity also emerges in KD, even 
when the distillation data are benign. Importantly, the effect is substantially weaker 
with non-chat KD. On Gemma, SORRY-Bench ASR increases by only \textbf{+3.14\%}, 
compared with \textbf{+35.32\%} under chat-template KD. The same pattern holds across 
model families: LLaMA shows a smaller HarmBench increase, while Qwen remains near 
baseline on HarmBench (\textbf{+0.12\%}) but degrades substantially under 
chat-template KD (\textbf{+13.12\%}).

These results suggest that using a non-chat template during KD reduces safety 
degradation under standard chat-template evaluation. However, non-chat KD does not 
fully preserve the original alignment. For example, LLaMA still shows a 
\textbf{+17.63\%} HarmBench ASR increase, indicating that distillation itself weakens 
refusal behavior. Although non-chat KD still improves utility over the base student 
model, its gains are smaller than those achieved by chat template KD. 


\subsection{The safety-utility trade-off.}

Given that KD is primarily used to improve task utility, it is important to examine 
whether template choice also affects this dimension. Our results reveal a clear 
trade-off: KD improves instruction-following under both template conditions, but larger 
utility gains coincide with larger ASR increases. As shown in 
Table~\ref{tab:result_prompt_template_safety_only}, chat-template KD achieves higher 
utility but also incurs the largest safety costs, whereas non-chat KD yields smaller 
utility gains while better preserving safety. Appendix~\ref{sec:temporal-chat-template} 
further shows that chat-template KD accelerates this vulnerability earlier in training.

One possible explanation is that the lower ASR of non-chat KD simply reflects weaker 
instruction-following rather than better safety preservation. However, this is 
inconsistent with the data: for LLaMA, non-chat KD achieves lower utility than SFT 
Chat (24.75 vs.\ 29.10) yet shows higher HarmBench ASR (29.81\% vs.\ 22.50\%). If 
reduced compliance explained the lower ASR, lower utility should correspond to lower 
ASR, which is not observed. Section~\ref{sec:mechanistic_analysis} further shows that 
non-chat KD better preserves refusal-related representations, suggesting deeper 
mechanistic differences between the two template conditions.

%% file: assets/table/result_prompt_template.tex
\begin{table*}[!h]
\centering
\small
\vspace{5pt}
\resizebox{\linewidth}{!}{%
\begin{tabular}{ll|l|cccc|c}
\toprule
\multirow{2}{*}{\textbf{Model}} 
& \multirow{2}{*}{\textbf{Method}}
& \multirow{2}{*}{\begin{tabular}{c}
\textbf{Training} \\
\textbf{Template}
\end{tabular}}
& \multicolumn{4}{c|}{\textbf{Safety (ASR\% $\downarrow$)}}
& \textbf{Utility (RL$\uparrow$) } \\
\cmidrule{4-8}
& & 
& \textbf{AdvBench} 
& \textbf{JBB} 
& \textbf{SORRY} 
& \textbf{HarmBench} 
& \textbf{AVG} \\
\midrule

\multirow{5}{*}{LLaMA-3.2-3B-Instruct}
& $\mathcal{M}_{S}$ 
& -
& 2.98 
& 7.25 
& 26.72 
& 12.18 
& 19.84 \\

\cmidrule{2-8}

& SFT 
& Non-chat
& 0.0096$_{\textcolor{purple}{-2.97}}$
& 0.04$_{\textcolor{teal}{-7.21}}$
& 27.12$_{\textcolor{purple}{+0.40}}$
& 12.19$_{\textcolor{purple}{+0.01}}$
& 24.27 \\

& SFT 
& Chat
& 0.04$_{\textcolor{purple}{-2.94}}$
& 0.07$_{\textcolor{purple}{-7.18}}$
& 28.71$_{\textcolor{purple}{+1.99}}$
& 22.50$_{\textcolor{purple}{+10.32}}$
& 29.10 \\

\cmidrule{2-8}

& KD 
& Non-chat
& 3.07$_{\textcolor{purple}{+0.09}}$
& 6.00$_{\textcolor{teal}{-1.25}}$
& 27.36$_{\textcolor{purple}{+0.64}}$
& 29.81$_{\textcolor{purple}{+17.63}}$
& 24.75 \\

& KD 
& Chat
& \textbf{4.99$_{\textcolor{purple}{+2.01}}$}
& \textbf{8.75$_{\textcolor{purple}{+1.50}}$}
& \textbf{30.86$_{\textcolor{purple}{+4.14}}$}
& \textbf{39.75$_{\textcolor{purple}{+27.57}}$}
& 29.03 \\

\midrule

\multirow{5}{*}{Gemma-2-2B-IT}
& $\mathcal{M}_{S}$ 
& -
& 0.00 
& 1.50 
& 16.04 
& 11.56 
& 22.35 \\

\cmidrule{2-8}

& SFT 
& Non-chat
& 0.00$_{\textcolor{purple}{+0.00}}$
& 0.015$_{\textcolor{teal}{-1.485}}$
& 28.71$_{\textcolor{purple}{+12.67}}$
&20.37$_{\textcolor{purple}{+8.81}}$
& 23.66 \\

& SFT 
& Chat
& 0.028$_{\textcolor{purple}{+0.028}}$
& 0.055$_{\textcolor{teal}{-1.445}}$
& 53.18$_{\textcolor{purple}{+37.14}}$
& 13.25$_{\textcolor{purple}{+1.69}}$
& 30.05 \\

\cmidrule{2-8}

& KD 
& Non-chat
& 0.00$_{\textcolor{purple}{+0.00}}$
& 1.50$_{\textcolor{purple}{+0.00}}$
& 19.18$_{\textcolor{purple}{+3.14}}$
& 14.25$_{\textcolor{purple}{+2.69}}$
& 23.63 \\

& KD 
& Chat
& \textbf{4.00$_{\textcolor{purple}{+4.00}}$}
& \textbf{6.50$_{\textcolor{purple}{+5.00}}$}
& \textbf{51.36$_{\textcolor{purple}{+35.32}}$}
& \textbf{19.56$_{\textcolor{purple}{+8.00}}$}
& 30.18 \\

\midrule

\multirow{5}{*}{Qwen-2.5-3B-Instruct}
& $\mathcal{M}_{S}$ 
& -
& 0
& 2.75
& 36.29
& 17.69
& 22.46 \\

\cmidrule{2-8}

& SFT 
& Non-chat
& 0.67$_{\textcolor{purple}{+0.67}}$
& 3.25$_{\textcolor{purple}{+0.50}}$
& 35.15$_{\textcolor{teal}{-1.14}}$
& 17.75$_{\textcolor{purple}{+0.06}}$
& 23.93 \\

& SFT 
& Chat
& 4.90$_{\textcolor{purple}{+4.90}}$
& 12.75$_{\textcolor{purple}{+10.00}}$
& 41.67$_{\textcolor{purple}{+5.38}}$
& 31.94$_{\textcolor{purple}{+14.25}}$
& 29.15 \\

\cmidrule{2-8}

& KD 
& Non-chat
& 1.06$_{\textcolor{purple}{+1.06}}$
&3.25$_{\textcolor{purple}{+0.50}}$
& 34.93$_{\textcolor{teal}{-1.36}}$ 
& 17.81$_{\textcolor{purple}{+0.12}}$
& 23.96 \\

& KD 
& Chat
& \textbf{5.87$_{\textcolor{purple}{+5.87}}$}
& \textbf{11$_{\textcolor{purple}{+8.25}}$}
& \textbf{40.99$_{\textcolor{purple}{+4.70}}$}
& \textbf{30.81$_{\textcolor{purple}{+13.12}}$}
& 28.95 \\

\bottomrule
\end{tabular}
}
\caption{\textbf{Student-model safety and utility under different prompt templates.}
We evaluate how prompt templates used during knowledge distillation affect the resulting student model's safety and utility. Safety is measured using Attack Success Rate (ASR\%), where lower values indicate safer behavior. Utility reports the average ROUGE-L score across four instruction-following benchmarks: Dolly, SelfInst, S-NI, and Vicuna. The subscripted deltas show the change relative to the corresponding student baseline ($\mathcal{M}_{S}$ ) within each model family.
Red deltas indicate increased ASR, corresponding to worse safety, while green deltas indicate reduced ASR, corresponding to improved safety. To emphasize the KD template effect, bold values are used only in the KD safety columns and denote the higher (less safe) ASR between chat and non-chat KD.}
\label{tab:result_prompt_template_safety_only}
\end{table*}

%% file: content/Analysis.tex
\section{Analysis}
\label{analysis_section}
The observed safety degradation under chat-template KD Prompts us to investigate three follow-up questions: first, whether this degradation is driven primarily by the student's training format or by the teacher's output distribution (\S\ref{sec:kd-template-crisscross}); second, whether this behavioral gap corresponds to changes in the model's internal refusal representations (\S\ref{sec:mechanistic_analysis}); and third, whether reducing the proportion of chat-template examples during training can moderate this effect (\S\ref{sec:prompt_mixing}).

\subsection{KD Prompt Template Drives the Degradation}
\label{sec:kd-template-crisscross}
\input{assets/table/criss_cross_evaluation}

To determine whether the safety degradation observed in Section~\ref{sec:result} is inherited from the teacher's output distribution or driven by the student's training format, we decouple the two templates during KD. As shown in Table~\ref{tab:crisscross_teacher_student_template_eval}, using the chat template only for the teacher does not meaningfully degrade safety, whereas using it only for the student causes a small but consistent regression across both model families. This suggests that the student-side training format plays the dominant role, rather than the teacher-side output format alone. We attribute this asymmetry to the nature of each side's influence: the student template directly shapes the input distribution over which gradients are computed, whereas the teacher template only determines the soft-target distribution observed by the student. As a result, the teacher-side template provides a weaker signal and does not directly alter the student's own representation learning.

The strongest degradation appears when both the student and teacher use the chat template. In this matched chat/chat setting, LLaMA HarmBench increases by $+27.57\%$ and Gemma SORRY-Bench increases by $+35.32\%$, far exceeding the changes observed when only one side uses the chat template. These results indicate that the regression is not simply inherited from the teacher, but emerges when the student learns from a teacher distribution generated under the same chat-based template. Decoupling either side removes most of the safety loss, showing that matched chat-template distillation is the primary source of the observed degradation. Since the student-side template emerges as the primary driver, we next examine whether this behavioral difference is also reflected in the model's internal representation.

\subsection{Mechanistic Analysis}
\label{sec:mechanistic_analysis}

\input{content/MechanisticAnalysis}

\input{assets/table/prompt_mixing_result}

\subsection{Template Mixing Provides Limited Mitigation}
\label{sec:prompt_mixing}

The preceding analyses show that using the chat template during KD drives behavioral safety degradation and shifts the model's internal refusal representation. We therefore examine whether this degradation depends on the amount of chat-template exposure during distillation. Rather than introducing additional safety-specific supervision~\cite{han2024wildguard, qi2023fine}, we isolate the role of prompt formatting by mixing chat  and non-chat formatted samples within the same benign utility dataset. This design allows us to test whether gradually reducing chat-template exposure can moderate safety degradation while keeping the training data content fixed and avoiding explicit safety supervision.

Specifically, we vary the proportion of chat template samples from 0 (pure non-chat) 
to 1 (pure chat), where intermediate values represent mixed training. A value of 0.5 chat-template proportion indicates that half of the training examples use the chat template, while the remaining half use the corresponding non-chat prompt format. As reported in 
Table~\ref{tab:combined_kd_results_prompt_mixing}, utility increases as this proportion 
increases: for LLaMA, AVG utility rises from 24.75 to roughly 29, and for Gemma from 
23.63 to roughly 30. However, these gains are small relative to the concurrent rise in ASR.

In contrast, safety is much more sensitive to the proportion of chat template samples. For LLaMA, 
SORRY-Bench increases from 27.36 to 30.86, while HarmBench rises sharply from 29.81 
to 39.75. A stronger trend appears for Gemma, where SORRY-Bench increases from 19.18 
to 51.36 and HarmBench from 14.25 to 19.56. Importantly, prompt mixing does not 
reduce ASR below the pure non-chat setting; it only attenuates degradation relative 
to full chat-template KD. These results suggest that prompt-template mixing minimally affects utility but substantially increases harmful-response susceptibility as chat-template proportions rise.

\subsection{Robustness to Prompt Formatting}
\label{token_ablation_cross_template_eval}

Our primary experiments use the standard chat template for all models, allowing controlled comparison across KD training formats. To assess whether the observed degradation depends on the inference format or on sensitivity to specific control tokens within the chat template, we perform two robustness analyses: cross-template evaluation, which changes the complete inference format, and token ablations, which selectively remove components of the chat template. 

\begin{table}[h]
\centering
\scriptsize
\setlength{\tabcolsep}{2pt}
\renewcommand{\arraystretch}{1.03}
\resizebox{\columnwidth}{!}{%
\begin{tabular}{llcccc}
\toprule
& & \multicolumn{2}{c}{\textbf{SORRY (ASR\% $\downarrow$)}}
& \multicolumn{2}{c}{\textbf{HarmBench(ASR\% $\downarrow$)}} \\
\cmidrule(lr){3-4}\cmidrule(lr){5-6}
\textbf{Model} & \textbf{Method} & \textbf{Chat} & \textbf{Non-chat} & \textbf{Chat} & \textbf{Non-chat} \\
\midrule

\multirow{3}{*}{LLaMA}
& Base
& 26.72 & 35.15
& 12.18 & 35.44 \\

& KD (Chat)
& 30.86$_{\tiny\textcolor{purple}{+4.14}}$
& 54.39$_{\tiny\textcolor{purple}{+19.24}}$
& 39.75$_{\tiny\textcolor{purple}{+27.57}}$
& 42.56$_{\tiny\textcolor{purple}{+7.12}}$ \\

& KD (Non-chat)
& 27.36$_{\tiny\textcolor{purple}{+0.64}}$
& 50.61$_{\tiny\textcolor{purple}{+15.46}}$
& 29.81$_{\tiny\textcolor{purple}{+17.63}}$
& 33.31$_{\tiny\textcolor{teal}{-2.13}}$ \\
\midrule

\multirow{3}{*}{Gemma}
& Base
& 16.04 & 19.92
& 11.56 & 17.75 \\

& KD (Chat)
& 51.36$_{\tiny\textcolor{purple}{+35.32}}$
& 71.51$_{\tiny\textcolor{purple}{+51.59}}$
& 19.56$_{\tiny\textcolor{purple}{+8.00}}$
& 27.56$_{\tiny\textcolor{purple}{+9.81}}$ \\

& KD (Non-chat)
& 19.18$_{\tiny\textcolor{purple}{+3.14}}$
& 40.99$_{\tiny\textcolor{purple}{+21.07}}$
& 14.25$_{\tiny\textcolor{purple}{+2.69}}$
& 34.81$_{\tiny\textcolor{purple}{+17.06}}$ \\
\midrule

\multirow{3}{*}{Qwen}
& Base
& 36.29 & 44.92
& 17.69 & 33.94 \\

& KD (Chat)
& 40.99$_{\tiny\textcolor{purple}{+4.70}}$
& 54.09$_{\tiny\textcolor{purple}{+9.17}}$
& 30.81$_{\tiny\textcolor{purple}{+13.12}}$
& 38.56$_{\tiny\textcolor{purple}{+4.62}}$ \\

& KD (Non-chat)
& 34.93$_{\tiny\textcolor{teal}{-1.36}}$
& 57.65$_{\tiny\textcolor{purple}{+12.73}}$
& 17.81$_{\tiny\textcolor{purple}{+0.12}}$
& 38.44$_{\tiny\textcolor{purple}{+4.50}}$ \\
\bottomrule
\end{tabular}%
}

\caption{\textbf{Cross-template safety evaluation.}Attack success rate (ASR\%, $\downarrow$) under chat and non-chat inference formats across the original instruction-tuned student (base) and distilled models. The subscripted deltas show the change relative to the corresponding student baseline.}
\label{tab:cross_template_eval}
\end{table}

\paragraph{Cross-template evaluation.} To test whether the observed degradation is caused by sensitivity to the inference template, we provide cross-template evaluation in HarmBench and SORRYBench safety dataset as shown in Table \ref{tab:cross_template_eval}.

For each model family, we compare the base and the distilled models under the same inference template. The degradation persists under non-chat evaluation, with chat template KD exhibits higher ASR than the corresponding base model across all three model families and both benchmarks. Although the magnitude of the degradation varies with the inference format, its direction remains consistent, indicating that the observed degradation in safety alignment is not limited to the standard chat template evaluation.

\begin{table}[t]
\centering
\scriptsize
\setlength{\tabcolsep}{3pt}
\resizebox{\columnwidth}{!}{%
\begin{tabular}{llccc}
\toprule
\textbf{Model} & \textbf{Training}
& \textbf{w/o BOS/EOT}
& \textbf{w/o Role}
& \textbf{Raw Query} \\
\midrule

\multirow{3}{*}{LLaMA}
& Base
& 16.38 & 22.31 & 35.69 \\

& KD-Non-chat
& 17.50$_{\tiny\textcolor{purple}{+1.12}}$
& 22.50$_{\tiny\textcolor{purple}{+0.19}}$
& 28.19$_{\tiny\textcolor{teal}{-7.50}}$ \\

& KD-Chat
& 25.75$_{\tiny\textcolor{purple}{+9.37}}$
& 24.75$_{\tiny\textcolor{purple}{+2.44}}$
& 39.81$_{\tiny\textcolor{purple}{+4.12}}$ \\
\midrule

\multirow{3}{*}{Gemma}
& Base
& 15.69 & 17.81 & 14.50 \\

& KD-Non-chat
& 19.00$_{\tiny\textcolor{purple}{+3.31}}$
& 22.94$_{\tiny\textcolor{purple}{+5.13}}$
& 16.06$_{\tiny\textcolor{purple}{+1.56}}$ \\

& KD-Chat
& 23.38$_{\tiny\textcolor{purple}{+7.69}}$
& 26.00$_{\tiny\textcolor{purple}{+8.19}}$
& 17.00$_{\tiny\textcolor{purple}{+2.50}}$ \\
\midrule

\multirow{3}{*}{Qwen}
& Base
& 32.44 & 48.75 & 32.13 \\

& KD-Non-chat
& 36.19$_{\tiny\textcolor{purple}{+3.75}}$
& 44.50$_{\tiny\textcolor{teal}{-4.25}}$
& 31.94$_{\tiny\textcolor{teal}{-0.19}}$ \\

& KD-Chat
& 38.44$_{\tiny\textcolor{purple}{+6.00}}$
& 45.75$_{\tiny\textcolor{teal}{-3.00}}$
& 32.75$_{\tiny\textcolor{purple}{+0.62}}$ \\
\bottomrule
\end{tabular}%
}

\caption{\textbf{Token-ablation safety evaluation.}
HarmBench ASR after removing chat-template components.
The subscripted deltas show the change relative to the corresponding student baseline.}
\label{tab:template_ablation_safety}
\end{table}
\paragraph{Token ablation.}

To examine whether the degradation is driven by specific control tokens in the chat template, we perform fine-grained HarmBench ablations across all three model families. We remove BOS/EOT tokens (\textit{w/o BOS/EOT}), which mark sequence or turn boundaries; remove role markers (\textit{w/o Role Tags}), which identify the user and assistant turns; and evaluate the harmful instruction alone without chat template tokens (\textit{Raw Query}) as shown in Table \ref{tab:template_ablation_safety}.

When BOS/EOT tokens are removed, KD-Chat remains less safe than the corresponding base model across all three families, with ASR increases of $+9.37$, $+7.69$, and $+6.00$ for LLaMA, Gemma, and Qwen, respectively. This  indicates that BOS/EOT tokens alone do not account for the increased ASR. The role-marker and raw-query ablations show more model-dependent behavior, as the base models themselves are highly sensitive to these input formats. Thus, individual template components can modulate the magnitude of the effect, but no single control token component consistently explains the degradation across model families.

%% file: assets/table/criss_cross_evaluation.tex
\begin{table}[!h]
\centering
\small

\vspace{3pt}
\resizebox{\columnwidth}{!}{%
\begin{tabular}{ll|cc|cc}
\toprule
\multirow{2}{*}{\textbf{Model}} &
\multirow{2}{*}{\textbf{Student}} &
\multicolumn{2}{c|}{\textbf{Teacher: Non-chat}} &
\multicolumn{2}{c}{\textbf{Teacher: Chat}} \\
\cmidrule(lr){3-4}\cmidrule(lr){5-6}
&
&
\textbf{SORRY $\downarrow$} &
\textbf{HarmBench $\downarrow$} &
\textbf{SORRY $\downarrow$} &
\textbf{HarmBench $\downarrow$} \\
\midrule

\multirow{2}{*}{LLaMA}
& Non-chat
& 27.36$_{\textcolor{purple}{+0.64}}$ 
& 29.81$_{\textcolor{purple}{+17.63}}$ 
& 26.36$_{\textcolor{teal}{-0.36}}$ 
& 11.44$_{\textcolor{teal}{-0.74}}$ \\
& Chat 
& 28.33$_{\textcolor{purple}{+1.61}}$ 
& 13.44$_{\textcolor{purple}{+1.26}}$ 
& 30.86$_{\textcolor{purple}{+4.14}}$ 
& 39.75$_{\textcolor{purple}{+27.57}}$ \\

\midrule

\multirow{2}{*}{Gemma}
& Non-chat 
& 19.18$_{\textcolor{purple}{+3.14}}$ 
& 14.25$_{\textcolor{purple}{+2.69}}$ 
& 16.90$_{\textcolor{purple}{+0.86}}$ 
& 12.50$_{\textcolor{purple}{+0.94}}$ \\
& Chat 
& 18.33$_{\textcolor{purple}{+2.29}}$ 
& 12.75$_{\textcolor{purple}{+1.19}}$ 
& 51.36$_{\textcolor{purple}{+35.32}}$ 
& 19.56$_{\textcolor{purple}{+8.00}}$ \\

\bottomrule
\end{tabular}%
}
\caption{Decoupling student and teacher templates during KD. Safety is measured using ASR (\%), where lower is better. The subscripted deltas show the change relative to the corresponding student baseline ($\mathcal{M}_{S}$).}
\label{tab:crisscross_teacher_student_template_eval}
\end{table}

%% file: content/MechanisticAnalysis.tex
As Section~\ref{sec:result} shows that the student-side template is the primary driver of safety degradation, we next examine whether this difference also appears in the model’s internal representations, and not just in output behavior, using the refusal direction~\citep{arditi2024refusal} as a diagnostic probe. Specifically, we analyze whether knowledge distillation changes the refusal geometry of the original instruction-tuned student model.

\paragraph{Setup.}
We probe the model's internal representation using the \textbf{refusal direction}~\citep{arditi2024refusal}, 
a linear axis in activation space estimated from residual stream activations (\texttt{resid\_pre}) 
that separates harmful from harmless prompts in the base student.
If distillation preserves this axis, the model retains its internal distinction between harmful and harmless inputs; if the axis changes or becomes weaker, the behavioral safety gap reflects a deeper representational shift. For each 
model and layer $\ell$, we estimate this direction from mean activation differences between harmful 
prompts (SORRY-Bench) and harmless prompts (Dolly-15k):

{\footnotesize
\begin{equation}
r_{\ell} =
\frac{
\mathbb{E}[h_{\ell}^{\text{harm}}] - \mathbb{E}[h_{\ell}^{\text{safe}}]
}{
\left\|
\mathbb{E}[h_{\ell}^{\text{harm}}] - \mathbb{E}[h_{\ell}^{\text{safe}}]
\right\|_2
}.
\end{equation}
}

We compute this direction for the base  student, the chat template distilled student, and the non-chat template 
distilled student. We then track two complementary properties after distillation: whether the refusal 
direction remains aligned with the base student (cosine similarity), and whether it still separates 
harmful from harmless activations (projection gap):

{\footnotesize
\begin{equation}
\mathrm{CosSim}_{\ell}(M, M_{S}) =
\frac{r_{\ell}^{M} \cdot r_{\ell}^{M_{S}}}
{\left\|r_{\ell}^{M}\right\|_2 \left\|r_{\ell}^{M_{S}}\right\|_2},
\end{equation}
\begin{equation}
\Delta_{\ell}^{M} =
\mathbb{E}_{x \in \mathcal{D}_{harm}}
\!\left[ h_{\ell}^{M}(x) \cdot r_{\ell}^{M} \right]
-
\mathbb{E}_{x \in \mathcal{D}_{safe}}
\!\left[ h_{\ell}^{M}(x) \cdot r_{\ell}^{M} \right].
\end{equation}
}

A cosine similarity close to 1.0 indicates that the refusal direction is preserved, while lower values 
indicate rotation away from the original direction. Furthermore, higher projection gaps indicate stronger internal 
separation between harmful and harmless prompts. We additionally apply PCA to last-token hidden states 
to visualize these shifts.  All models are evaluated using the same chat-template inference format so that differences reflect the effect of distillation rather than inference-time formatting.

\paragraph{Findings.}

\input{assets/table/mechanistic_analysis}

As shown in Table~\ref{tab:rep_similarity}, chat-template KD causes larger changes in refusal geometry than non-chat-template KD across Gemma, LLaMA, and Qwen. Across all three model families, non-chat distillation better preserves the base model's representation structure, while chat-template distillation induces a larger representational shift and reduces the projection gap. The chat-template distilled models show lower cosine similarity to the original instruction-tuned student's refusal direction, indicating that the refusal direction is less preserved after chat-template KD. They also show a weaker projection gap, suggesting that harmful and harmless prompts become less separable along the refusal direction. This pattern is consistent with the behavioral safety results: the same condition that produces higher attack success rate also produces larger shifts in refusal-related representations.

\begin{figure}[!t]
    \centering
    \includegraphics[width=0.9\linewidth]{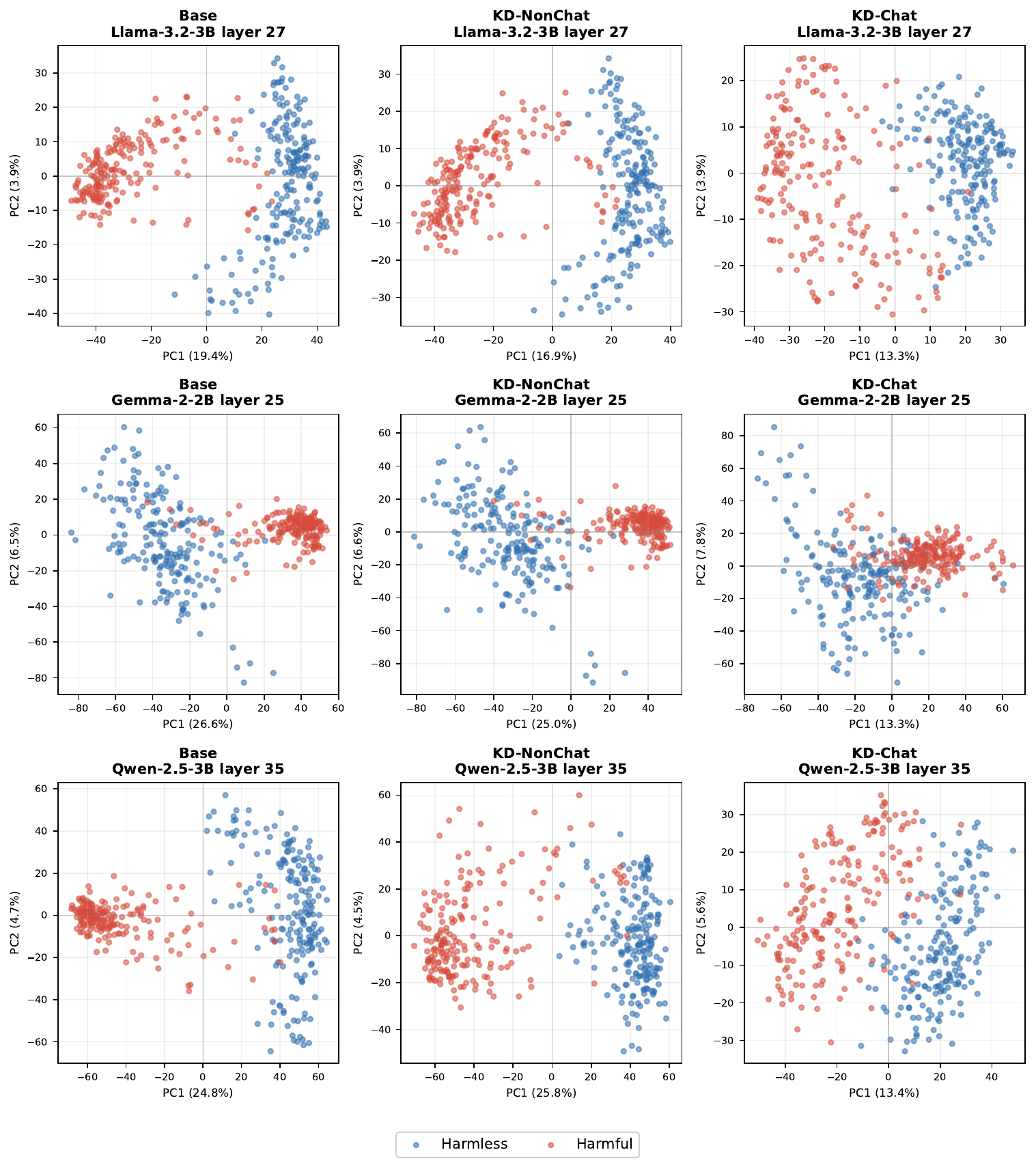}
  \caption{\textbf{Final-layer representation shifts under different prompt templates.} PCA projections of final-layer hidden states across three models show that non-chat template KD remains closer to the base instruct-tuned student, whereas chat-template KD produces a larger shift in representation space.}
    \label{fig:llama_all_pca_projections}
\end{figure}

In contrast, non-chat template KD better preserves both the direction and separability of the original refusal geometry. This suggests that non-chat template distillation interferes less with internal representations associated with refusal behavior. The PCA visualization in Figure~\ref{fig:llama_all_pca_projections} provides a complementary view of this effect, showing that chat template KD leads to greater changes in the harmful-vs-harmless representation structure. Overall, these findings suggest that the safety degradation caused by chat template KD is not only a surface-level decoding effect, but is also reflected in intermediate representations.

We further perform small-scale causal interventions on the learned refusal
direction as a diagnostic test of the correlational findings. The results
are reported in Appendix~\ref{prelims:causal_intervention}, with an important
caveat that the LLaMA ablation can impair output coherence and therefore its
lower ASR should not be interpreted as improved safety.

%% file: assets/table/mechanistic_analysis.tex
\begin{table}[!h]
\centering
\resizebox{\linewidth}{!}{
\begin{tabular}{llcc}
\toprule
\textbf{Model} & \textbf{Method} & 
\begin{tabular}[c]{@{}c@{}}
\textbf{Cosine Similarity} \\
\textbf{to Base $\uparrow$}
\end{tabular} &
\begin{tabular}[c]{@{}c@{}}
\textbf{Projection} \\
\textbf{Gap $\uparrow$}
\end{tabular} \\
\midrule
\multirow{3}{*}{Gemma}
& Base        & ---  & 368.78 \\
& KD (Chat)     & 0.53 & 175.60 \\
& KD (Non-chat) & 0.99 & 355.76 \\
\midrule
\multirow{3}{*}{LLaMA}
& Base        & ---  & 20.38 \\
& KD (Chat)     & 0.75 & 18.48 \\
& KD (Non-chat)  & 0.97 & 20.87 \\
\midrule
\multirow{3}{*}{Qwen}
& Base        & ---  & 118.45 \\
& KD (Chat)     & 0.54 & 83.21 \\
& KD (Non-chat)  & 0.99 & 110.69 \\
\bottomrule
\end{tabular}
}
\caption{\textbf{Representation similarity and refusal-projection gap across model families.}
Cosine similarity is measured with respect to the refusal direction of the original instruction-tuned student, and projection gap measures the harmful--harmless separation along this direction.}
\label{tab:rep_similarity}
\end{table}

%% file: assets/table/prompt_mixing_result.tex
\begin{table}[!h]
\centering
\small

\vspace{5pt}
\resizebox{\linewidth}{!}{%
\begin{tabular}{lc|ccc}
\toprule
\multirow{2}{*}{\textbf{Model}}
& \multirow{2}{*}{\textbf{Chat Prop.}}
& \multicolumn{2}{c|}{\textbf{Safety (ASR\% $\downarrow$)}}
& \textbf{Utility (RL$\uparrow$)} \\
\cmidrule{3-5}
& 
& \textbf{SORRY}
& \textbf{HarmBench}
& \textbf{AVG} \\
\midrule

\multirow{5}{*}{LLaMA}
& $\mathcal{M}_{S}$
& 26.72
& 12.18
& 18.77 \\

\cmidrule{2-5}

& 0.0
& 27.36$_{\textcolor{purple}{+0.64}}$
& 29.81$_{\textcolor{purple}{+17.63}}$
& 24.75 \\

& 0.5
& 28.18$_{\textcolor{purple}{+1.46}}$
& 23.06$_{\textcolor{purple}{+10.88}}$
& 28.62 \\

& 0.8
& 29.85$_{\textcolor{purple}{+3.13}}$
& 24.50$_{\textcolor{purple}{+12.32}}$
& 28.74 \\

& 1.0
& 30.86$_{\textcolor{purple}{+4.14}}$
& 39.75$_{\textcolor{purple}{+27.57}}$
& 29.03 \\

\midrule

\multirow{5}{*}{Gemma}
& $\mathcal{M}_{S}$
& 16.04
& 11.56
& 22.35 \\

\cmidrule{2-5}

& 0.0
& 19.18$_{\textcolor{purple}{+3.14}}$
& 14.25$_{\textcolor{purple}{+2.69}}$
& 23.63 \\

& 0.5
& 38.94$_{\textcolor{purple}{+22.90}}$
& 16.88$_{\textcolor{purple}{+5.32}}$
& 29.60 \\

& 0.8
& 44.39$_{\textcolor{purple}{+28.35}}$
& 18.50$_{\textcolor{purple}{+6.94}}$
& 30.24 \\

& 1.0
& 51.36$_{\textcolor{purple}{+35.32}}$
& 19.56$_{\textcolor{purple}{+8.00}}$
& 30.18 \\

\bottomrule
\end{tabular}
}
\caption{\textbf{KD safety and utility under different chat template proportions.}
Chat Prop. denotes the proportion of chat-template examples used during KD, where 0.0 is pure non-chat and 1.0 is pure chat template.}

\label{tab:combined_kd_results_prompt_mixing}
\end{table}

%% file: content/Conclusion.tex
\section{Discussion}

The above analyses in section \ref{analysis_section} show that chat template KD consistently leads to safety degradation than non-chat KD, is accompanied by changes in refusal-related representations, increases with chat template exposure, and remains evident across alternative inference formats and token ablations. Therefore, these findings show that safety degradation during knowledge distillation is not only a consequence of optimizing on benign task data, but is also strongly shaped by the prompt template used during training. In particular, chat template KD produces faster and larger increases in ASR than non-chat  KD, indicating that prompt formatting can amplify the erosion of safety alignment. These results suggest that prompt formatting during KD affects safety beyond a simple evaluation-template artifact.

We hypothesize that this occurs because chat templates expose structural role tokens and conversational transitions, such as user-assistant boundaries, that are closely tied to the model's learned refusal behavior. During KD, repeatedly optimizing on benign chat-formatted responses may overwrite or weaken these safety-relevant associations, thereby disrupting the conditional behavior needed for refusal. This interpretation is consistent with our representation-level analysis, where chat-template KD produces larger shifts in refusal-related geometry compared to non-chat-template KD.

A practical implication of these findings is that prompt templates should be decoupled 
between training and deployment when safety is the primary concern. Specifically, our 
results suggest using a non-chat template during KD training while still deploying 
through the standard chat interface, i.e., \textit{non-chat training} $\rightarrow$ 
\textit{chat inference}. However, this training--deployment asymmetry is unusual in 
practice, and its robustness beyond benign instruction-following data and our benchmark 
settings remains to be verified. Thus, prompt templates are not merely a syntactic 
implementation choice, but an important factor in alignment stability.

\section{Conclusion}

In this work, we show that knowledge distillation (KD) on benign tasks can erode the safety alignment of student models. Across model families, KD improves task utility but also increases harmful-compliance vulnerability, suggesting that optimizing for instruction-following performance can conflict with preserving the refusal behavior already present in aligned student models. Our findings further identify prompt templates as a key training-time factor that amplifies this degradation. More broadly, these results raise an important question for future work: whether fine-tuning and distillation methods can preserve refusal geometry without relying on explicit safety supervision, enabling student models to improve utility while maintaining the safety behavior of their base instruct-tuned models.

%% file: content/Limitations.tex
\section*{Limitations}

While our study provides evidence that prompt templates play an important role in 
safety degradation during knowledge distillation, three limitations remain.

\paragraph{Dataset.} First, KD is performed exclusively on benign 
instruction-following data, leaving open whether similar template-driven 
degradation emerges in domain-specific settings such as code, math, or 
reasoning. These domains differ in how heavily they rely on structural 
tokens: code contains formatting cues that may interact with chat-template 
control tokens in distinct ways, whereas math and reasoning data are 
largely plain text. We restrict our study to the benign instruction-following 
setting in order to isolate the effect of prompt formatting from 
domain-induced distributional shifts; extending the analysis to 
structurally richer domains is a natural next step that our pipeline 
directly supports.

\paragraph{Beyond LoRA.} Second, all experiments use LoRA-based distillation rather than full-parameter fine-tuning. Prior work suggests that LoRA can produce behavioral and representational changes similar to full fine-tuning~\citep{hu2022lora}. Therefore, we expect the template-driven safety degradation observed in our experiments to also appear under full-parameter distillation, although the magnitude may differ. A direct comparison between LoRA and full-parameter training is an important direction for future work.

\paragraph{Evaluator Bias.} Safety evaluation uses automated judges (LLaMA-Guard-8B for AdvBench and JailbreakBench, a fine-tuned Mistral judge for SORRY-Bench, and LLaMA-2-13B-cls for HarmBench). These judges may not fully capture all forms of harmful content. However, each benchmark uses a fixed judge, so comparisons within a benchmark remain consistent even if absolute ASR values differ. A direct comparison across judges is left for future work.

\section*{Ethical Considerations}

This work investigates conditions under which knowledge distillation 
on benign downstream tasks can erode the safety alignment of LLMs. While these findings could in principle be exploited to bypass refusal 
behavior, our research primary intent is to highlight underexplored vulnerability in the KD pipeline. We aim to inform the development of more robust safety-preserving distillation techniques. 
All experiments use publicly available 
datasets and models, and no human subjects were involved. The safety evaluation datasets contain harmful or offensive content by design and were used solely for research and evaluation. 

%% file: content/Acknowledgements.tex
\section*{Acknowledgments}

Research was sponsored by the Army Research Laboratory and was accomplished under Cooperative Agreement Number W911NF-23-2-0224. The views and conclusions contained in this document are those of the authors and should not be interpreted as representing the official policies, either expressed or implied, of the Army Research Laboratory or the U.S. Government. The U.S. Government is authorized to reproduce and distribute reprints for Government purposes notwithstanding any copyright notation herein.

%% file: content/Appendix.tex
\section{Training Details}
\subsection{Hyperparameter details}
\label{sec:training_settings}

\input{assets/table/hyperparameter}

 Table~\ref{tab:hyperparameter} summarizes the hyperparameter settings used for student model distillation. We use Brain Floating Point (BF16) mixed precision to improve training throughput while maintaining numerical stability. During instruction-following evaluation, responses are generated with a decoding temperature of $t=1.0$ and nucleus sampling parameter $top\_p=0.7$.

\subsection{Dataset Details}
\label{sec:dataset_details}

\input{assets/table/data_stat}
In Table~\ref{tab:data_stats}, we provide the details of training and evaluation datasets used for downstream instruction-following tasks and the benchmarks employed for safety evaluation.

\subsection{Prompt Template}
\input{assets/table/prompt_template}

An example of each prompt template used during distillation is shown in Table~\ref{tab:prompt-template}.

\subsection{Safety Evaluation}
\label{safety_eval}
In this section, we summarize the safety benchmarks employed in to assess both the pre-exisiting student models safety and their safety alignment after distillation process.

\textbf{AdvBench:} We evaluate model safety on AdvBench, containing 520 harmful behaviors \cite{zou2023universal}. Following WalledEval, we use their curated harmful prompts 
\footnote{\raggedright AdvBench dataset: \href{https://huggingface.co/datasets/walledai/AdvBench}{\nolinkurl{walledai/AdvBench}}. \par}  and prompt to the models. Then, the responses are assessed using \texttt{LLaMA-3-Guard-8B} \citep{grattafiori2024llama}. 

\textbf{JailbreakBench:} We further evaluate refusal robustness using 200 harmful prompts, from JailbreakBench \cite{chao2024jailbreakbench}. The evaluation was done by the same model as AdvBench. The reported score of AdvBench and JailbreakBench, in \ref{tab:result_prompt_template_safety_only} is the average of two seeds. 

\textbf{SORRY-Bench:} SORRY-Bench comprises a 44-class safety taxonomy with 440 base prompts and uses a fine-tuned Mistral-7B-Instruct-v0.2 model as an automated judge for safety refusal.\footnote{\raggedright SORRY-Bench judge model: \href{https://huggingface.co/SORRY-Bench/ft-mistral-7b-instruct-v0.2-SORRY-Bench-202406}{\nolinkurl{SORRY-Bench/ft-mistral-7b-instruct-v0.2-SORRY-Bench-202406}}. \par}

 We evaluate models on the base prompts and the fulfillment rate in the benchmark is reported here as ASR, defined as the proportion of responses that comply with unsafe instructions. Higher ASR indicates weaker safety, while lower ASR reflects stronger refusal. Evaluations use default decoding settings ($t=0.7$, $top\_p=1.0$, max tokens $=1024$), and reported results are averaged over three random seeds in Table~\ref{tab:result_prompt_template_safety_only}).

\textbf{HarmBench:} To evaluate the generalization of safety performance, we additionally assess models on HarmBench. Although HarmBench is primarily designed for automated red teaming and evaluating LLM attacks and defenses, we used it to measure the safety of models based on ability to elicit harmful behavior. HarmBench defines four behavior categories—standard, copyright, contextual, and multimodal—and we use the test split comprising 320 behaviors. For each behavior, we generate five test cases using \texttt{Mixtral-8x7B-Instruct-v0.1}\footnote{\url{https://huggingface.co/mistralai/Mixtral-8x7B-Instruct-v0.1}} Safety is evaluated using ASR, where lower values indicate safer models. We employ the provided \texttt{LLaMA-2-13b-cls} classifier as the automated judge\footnote{\raggedright HarmBench classifier: \href{https://huggingface.co/cais/HarmBench-LLaMA-2-13b-cls}{\nolinkurl{cais/HarmBench-LLaMA-2-13b-cls}}. \par}.
 All reported results correspond to zero-shot prompts generated with HarmBench’s default configuration.

\section{Safety-Utility}
\input{assets/table/safety_utility_prompt_template}
Table~\ref{tab:result_prompt_template_utility} presents a detailed breakdown of utility across the instruction-following evaluation datasets. All results are obtained using a decoding temperature of $t=1.0$ and $top\_p=0.7$.

\section{Generalization to other distillation objective}
\label{sec:other_distillation_objective}
\input{assets/table/other_distillation_objective}

The effect of prompt templates generalizes beyond the standard KD objective. As shown in Table~\ref{tab:other_distillation_objective}, chat-template distillation consistently produces higher ASR than non-chat distillation across RKL and FKL+RKL, while also yielding higher average utility. This pattern holds across all three models. For example, under RKL, HarmBench ASR increases from 12.56 to 26.63 for LLaMA and from 18.38 to 29.38 for Qwen when moving from non-chat to chat distillation. Gemma shows the same trend on SORRY-Bench, increasing from 18.86 to 41.29. These results indicate that template-driven safety degradation is robust to changes in the distillation objective.

\section{Teacher Model Performance}
\input{assets/table/teacher_student_sft_template}
Table~\ref{tab:abl_sft_teacher_student} reports supervised fine-tuning under both prompt templates on teacher and student checkpoints. We observed similar pattern of chat-template SFT consistently degrading safety across both model families and both scales similar to \cite{lyu2024keeping}, while non-chat SFT leaves the baseline largely intact (e.g., Gemma 9B SORRY: $+34.31$ vs.\ $+0.53$; LLaMA 8B HarmBench: $+7.00$ vs.\ $-0.06$). 
The effect is therefore not a small-model artifact nor specific to distillation. 


\section{Chat templates accelerate vulnerability.} 
\label{sec:temporal-chat-template}
\begin{figure}[h]
    \centering
    \includegraphics[width=1\linewidth]{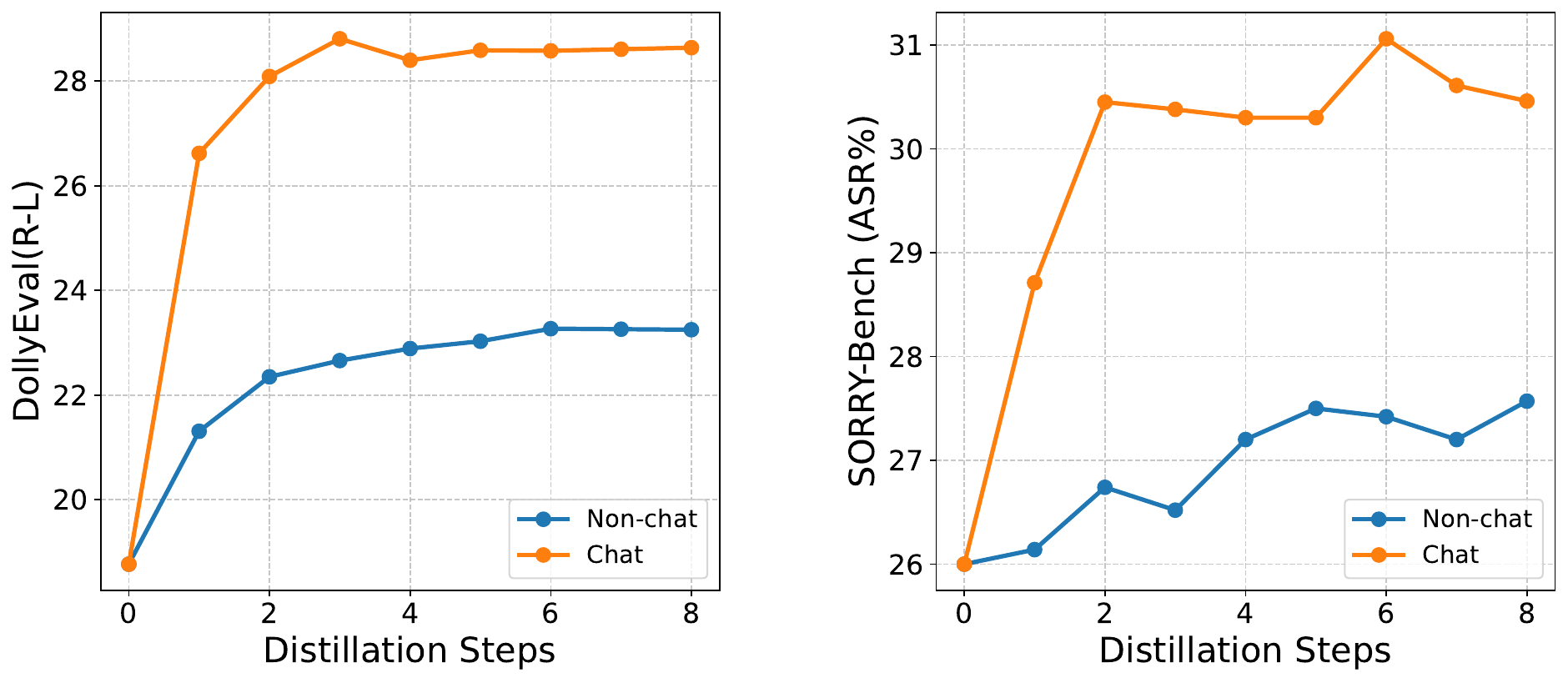}
    \caption{\textbf{Temporal dynamics of safety and utility.} Chat (\textcolor{orange}{orange}) triggers rapid utility and vulnerability spikes. Non-Chat (\textcolor{cyan}{blue}) follows a gradual trajectory, slowing safety erosion in student LLaMA model}
    \label{fig:utility_safety_comparison}
\end{figure}

To further examine the effect of prompt templates, we track task utility, measured by ROUGE-L, and safety degradation, measured by ASR, across distillation steps. As shown in Figure~\ref{fig:utility_safety_comparison}, chat-template KD produces a sharper early increase in both utility and ASR. This indicates that the chat template improves instruction-following performance quickly, but also accelerates the loss of safety behavior. In contrast, non-chat-template KD shows a more gradual trajectory, with slower increases in ASR while utility improves steadily. These results provide additional evidence that chat templates amplify safety degradation during distillation, whereas non-chat templates lead to a more stable training trajectory.

\begin{table}[h]
    \centering
    \scriptsize
    \setlength{\tabcolsep}{4pt}
    \resizebox{\columnwidth}{!}{%
    \begin{tabular}{lcc}
        \toprule
        \textbf{Model}
        & \textbf{Ablate KD (Chat)}
        & \textbf{Add to KD (Non-Chat)} \\
        \midrule
        LLaMA
        & 35.0 $\rightarrow$ 20.0
        & 15.0 $\rightarrow$ 5.0 \\
        Gemma
        & 25.0 $\rightarrow$ 62.5
        & 7.5 $\rightarrow$ 0.0 \\
        Qwen
        & 20.0 $\rightarrow$ 55.0
        & 12.5 $\rightarrow$ 0.0 \\
        \bottomrule
    \end{tabular}%
    }

    \caption{\textbf{Preliminary refusal-direction interventions.}
    ASR (\%) before and after intervention on 40 harmful SORRY-Bench prompts.
    ``Ablate KD (Chat)'' removes the KD-Chat refusal direction, while
    ``Add to KD (Non-Chat)'' adds the base model's refusal direction.}
    \label{tab:causal_interventions}
\end{table}

\section{Preliminary causal intervention on refusal direction.}
\label{prelims:causal_intervention}

To complement the correlational analysis in \S\ref{sec:mechanistic_analysis}, we perform small-scale interventions on the difference-in-means refusal direction. For each model family, we evaluate 40 harmful prompts from SORRY-Bench using the same judge as in the main evaluation. For KD-Chat, we ablate the learned refusal direction from the model activations to test whether weakening this direction increases unsafe behavior. For KD Non-Chat, we add the base model's refusal direction to test whether restoring this direction improves safety. The full results are reported in Table~\ref{tab:causal_interventions}.

When we ablate the refusal direction, ASR increases for Gemma and Qwen, while adding the base model's refusal direction to KD-Non-Chat reduces ASR across all three model families. These results provide preliminary interventional evidence that modifying the refusal direction can influence safety behavior.

For LLaMA student model, ablating the refusal direction decreases ASR, but this does not indicate improved safety. After the intervention, 9/40 outputs become repetitive or off-topic, compared with 1/40 before intervention. These incoherent outputs are scored as non-compliant by the judge, the measured ASR is mechanically reduced. We therefore attribute the lower ASR in this setting to degraded output coherence rather than improved refusal behavior.

\section{Full token ablations comparison}
Table~\ref{app:full_template_ablation_safety} provides the full HarmBench comparison across inference formats and token ablations. We report results under the standard chat template, after removing BOS/EOT tokens, after removing role tags, under the non-chat template, and finally using the raw harmful query without chat-template formatting. This ordering progressively removes chat-specific structure and allows us to compare how safety behavior changes as different components of the inference format are removed. Overall, KD-Chat remains less safe than the corresponding base model under several ablated settings, especially after removing BOS/EOT tokens, while the role-tag and raw-query conditions are more model dependent. These results suggest that no single chat-template component fully accounts for the degradation, although inference formatting can substantially modulate its magnitude.

\section{Full Token Ablation Comparison}

Table~\ref{app:full_template_ablation_safety} provides the full HarmBench comparison across inference formats and token ablations. We report results under the standard chat template, after removing BOS/EOT tokens, after removing role tags, under the non-chat template, and finally using the raw harmful query without chat-template formatting. This ordering progressively removes chat-specific structure and allows us to compare how safety behavior changes as different components of the inference format are removed. Overall, KD-Chat remains less safe than the corresponding base model under several ablated settings, especially after removing BOS/EOT tokens, while the role-tag and raw-query conditions are more model dependent. These results suggest that no single chat-template component fully accounts for the degradation, although inference formatting can substantially modulate its magnitude.

\begin{table*}[htbp]
\centering
\small
\setlength{\tabcolsep}{5pt}
\begin{tabular}{llccccc}
\toprule
\textbf{Model} & \textbf{Method}
& \multicolumn{5}{c}{\textbf{Safety (ASR\% $\downarrow$)}} \\
\cmidrule(lr){3-7}
& & \textbf{Chat Eval}
& \textbf{w/o BOS/EOT}
& \textbf{w/o Role Tags}
& \textbf{Non-chat Eval}
& \textbf{Raw Query} \\
\midrule

\multirow{3}{*}{LLaMA}
& Base
& 12.18
& 16.38
& 22.31
& 35.44
& 35.69 \\

& KD-Non-chat
& 29.81$_{\tiny\textcolor{purple}{+17.63}}$
& 17.50$_{\tiny\textcolor{purple}{+1.12}}$
& 22.50$_{\tiny\textcolor{purple}{+0.19}}$
& 33.31$_{\tiny\textcolor{teal}{-2.13}}$
& 28.19$_{\tiny\textcolor{teal}{-7.50}}$ \\

& KD-Chat
& 39.75$_{\tiny\textcolor{purple}{+27.57}}$
& 25.75$_{\tiny\textcolor{purple}{+9.37}}$
& 24.75$_{\tiny\textcolor{purple}{+2.44}}$
& 42.56$_{\tiny\textcolor{purple}{+7.12}}$
& 39.81$_{\tiny\textcolor{purple}{+4.12}}$ \\
\midrule

\multirow{3}{*}{Gemma}
& Base
& 11.56
& 15.69
& 17.81
& 17.75
& 14.50 \\

& KD-Non-chat
& 14.25$_{\tiny\textcolor{purple}{+2.69}}$
& 19.00$_{\tiny\textcolor{purple}{+3.31}}$
& 22.94$_{\tiny\textcolor{purple}{+5.13}}$
& 34.81$_{\tiny\textcolor{purple}{+17.06}}$
& 16.06$_{\tiny\textcolor{purple}{+1.56}}$ \\

& KD-Chat
& 19.56$_{\tiny\textcolor{purple}{+8.00}}$
& 23.38$_{\tiny\textcolor{purple}{+7.69}}$
& 26.00$_{\tiny\textcolor{purple}{+8.19}}$
& 27.56$_{\tiny\textcolor{purple}{+9.81}}$
& 17.00$_{\tiny\textcolor{purple}{+2.50}}$ \\
\midrule

\multirow{3}{*}{Qwen}
& Base
& 17.69
& 32.44
& 48.75
& 33.94
& 32.13 \\

& KD-Non-chat
& 17.81$_{\tiny\textcolor{purple}{+0.12}}$
& 36.19$_{\tiny\textcolor{purple}{+3.75}}$
& 44.50$_{\tiny\textcolor{teal}{-4.25}}$
& 38.44$_{\tiny\textcolor{purple}{+4.50}}$
& 31.94$_{\tiny\textcolor{teal}{-0.19}}$ \\

& KD-Chat
& 30.81$_{\tiny\textcolor{purple}{+13.12}}$
& 38.44$_{\tiny\textcolor{purple}{+6.00}}$
& 45.75$_{\tiny\textcolor{teal}{-3.00}}$
& 38.56$_{\tiny\textcolor{purple}{+4.62}}$
& 32.75$_{\tiny\textcolor{purple}{+0.62}}$ \\
\bottomrule
\end{tabular}

\caption{\textbf{Safety under inference-template and token ablations.}
HarmBench attack success rate (ASR\%) for models with training settings and inference configurations. The subscripted values report the change relative to the corresponding base model under the same evaluation configuration.}
\label{app:full_template_ablation_safety}
\end{table*}

\section{Use of AI Assistants}
We used AI assistants to help polish the text and debug code. All content, ideas, and analyses presented in this paper remain the sole responsibility of the authors.

%% file: assets/table/hyperparameter.tex
\begin{table}[h]
    \centering
    \scriptsize
    \setlength{\tabcolsep}{3pt}
    \renewcommand{\arraystretch}{1.15}
    
    \resizebox{\columnwidth}{!}{
    \begin{tabular}{l|c}
    \toprule
    \textbf{Hyperparameter} 
    & \textbf{Instruction} \\
    \midrule
    Effective Batch Size  & 64 \\
    Initial LR & $2.0 \times 10^{-5}$ \\
    LR Decay Style & Cosine \\
    Optimizer & AdamW \\
    Weight Decay & 0.01 \\
    Parallel Strategy & DDP \\
    Precision & BF16 \\
    Epochs & 8 \\
    Max Prompt Length & 512 \\
    Max Sequence Length & 1024 \\
    Evaluation Decoding 
    & $t=1.0$, $top\_p=0.7$ \\
    \bottomrule
    \end{tabular}
    }
    \caption{\textbf{Hyperparameter settings.} Dataset-wise training and evaluation configurations used for student model distillation.}
    \label{tab:hyperparameter}
\end{table}

%% file: assets/table/data_stat.tex
\begin{table}[h!]
\centering
\resizebox{\linewidth}{!}{
\begin{tabular}{llcc}
\toprule
\textbf{Type} & \textbf{Name} & \textbf{\# Train} & \textbf{\# Test} \\
\midrule
\multirow{4}{*}{Instruction Following}
& Dolly      & 11435 & 500 \\
& Self-Instruct & --   & 242 \\
& Vicuna     & --   & 80  \\
& S-NI       & --   & 1694 \\
\midrule
\multirow{4}{*}{Safety Evaluation}
& AdvBench      & -- & 520 \\
& JailbreakBench & --   & 200 \\
& SORRY-Bench     & --   & 400  \\
& HarmBench       & --   & 1600 \\
\bottomrule
\end{tabular}
}
\caption{Data statistics of training and evaluation data.}
\label{tab:data_stats}
\end{table}

%% file: assets/table/prompt_template.tex
\begin{table*}[!t]
    \small
    \centering
    
    \begin{tabularx}{\linewidth}{p{45pt}|p{100pt}|X}
        \textbf{Name} & \textbf{Prompt Template} & \textbf{Example Query} \\
        \hline
        Non-chat &
        \texttt{Instruction: <query> \newline Response: <output>} &
        Below is an instruction that describes a task. Write a response that appropriately completes the request. 
        \#\#\#\ Instruction: What percussion instruments are easy to learn? 
        \#\#\#\ Response: 
        \\
        \hline
        Llama (Chat) &
        \raggedright \texttt{<|system|>... \newline <|user|> <query> \newline <|assistant|>} &
        \sloppy \texttt{ <|begin\_of\_text|>  <|start\_header\_id|>system<|end\_header\_id|>\newline \{system prompt\}<|eot\_id|><|start\_header\_id|>user<|end\_header\_id|>} What percussion instruments are easy to learn? \texttt{<|eot\_id|><|start\_header\_id|>assistant<|end\_header\_id|>}
       \\
         \hline
        Gemma (Chat) &
        \raggedright \texttt{<bos><start\_of\_turn>user \newline <query><end\_of\_turn> \newline <start\_of\_turn>model} &
        \sloppy \texttt{<bos><start\_of\_turn>user} What percussion instruments are easy to learn? \texttt{<end\_of\_turn><start\_of\_turn>model}
       \\
         \hline
        Qwen (Chat) &
        \raggedright \texttt{<|im\_start|>system...\newline<|im\_end|> \newline <|im\_start|>user \newline <query><|im\_end|> \newline <|im\_start|>assistant} &
        \sloppy \texttt{<|im\_start|>system} You are Qwen, created by Alibaba Cloud. You are a helpful assistant. \texttt{<|im\_end|><|im\_start|>user} What percussion instruments are easy to learn? \texttt{<|im\_end|><|im\_start|>assistant}
       \\
         \hline
    \end{tabularx}
    \caption{Example of Prompt templates used for LLaMA, Gemma, and Qwen family. The non-chat is common to all the models whereas based on the model family chat template varies.}
    \label{tab:prompt-template}
\end{table*}

%% file: assets/table/safety_utility_prompt_template.tex
\begin{table*}[!t]
    \centering
    \small
    \vspace{5pt}
    \resizebox{\linewidth}{!}{%
        \begin{tabular}{lrl|cccc|c|c}
        \toprule
        \multirow{2}{*}{\textbf{Model}} 
        & \multirow{2}{*}{\textbf{\#Params}} 
        & \multirow{2}{*}{\textbf{Method}}
        & \multicolumn{4}{c|}{\textbf{Utility (R-L$\uparrow$)}} 
        & \textbf{Avg}
        & \textbf{Safety (ASR$\downarrow$}) \\
        \cmidrule{4-9}
        & &  
        & \textbf{Dolly} 
        & \textbf{SelfInst} 
        & \textbf{Vicuna} 
        & \textbf{S-NI} 
        & \textbf{R-L} 
        & \textbf{SORRY} \\ 
        \midrule

        \multirow{6}{*}{LLaMA}
        & 8B 
        & $\mathcal{M}_{T}$ 
        & 30.54 
        & 26.59 
        & 22.61 
        & 42.00 
        & 30.44 
        & -- \\ 

        \cmidrule{2-9}

        & \multirow{7}{*}{3B}
        & $\mathcal{M}_{S}$ 
        & 18.77 
        & 13.21 
        & 24.19 
        & 23.18 
        & 19.84 
        & 26.72 \\

        & 
        & SFT (Non-Chat) 
        & 22.36 
        & 16.13 
        & 25.33 
        & 33.27 
        & 24.27 
        & 27.12 \\

        & 
        & SFT (Chat) 
        & 29.37 
        & 24.65 
        & 22.68 
        & 39.68 
        & 29.10 
        & 28.71 \\

        & 
        & KD (Non-Chat) 
        & 23.36 
        & 16.45 
        & 25.22 
        & 33.99 
        & 24.75 
        & 27.36 \\

        & 
        & KD (Chat) 
        & 28.88 
        & 24.30 
        & 22.96 
        & 39.98 
        & 29.03 
        & \textbf{30.86} \\



        \midrule

        \multirow{6}{*}{Gemma}
        & 9B 
        & $\mathcal{M}_{T}$ 
        & 30.39 
        & 29.39 
        & 21.17 
        & 45.07 
        & 31.51 
        & -- \\ 

        \cmidrule{2-9}

        & \multirow{7}{*}{2B}
        & $\mathcal{M}_{S}$ 
        & 18.36 
        & 13.93 
        & 22.66 
        & 34.44 
        & 22.35 
        & 16.04 \\

        & 
        & SFT (Non-Chat) 
        & 18.19 
        & 15.26 
        & 22.81 
        & 38.39 
        & 23.66 
        & 17.80 \\

        & 
        & SFT (Chat) 
        & 28.78 
        & 27.66 
        & 20.97 
        & 42.78 
        & 30.05 
        & 53.18 \\

        & 
        & KD (Non-Chat) 
        & 18.52 
        & 15.04 
        & 22.85 
        & 38.10 
        & 23.63 
        & 19.18 \\

        & 
        & KD (Chat) 
        & 28.57 
        & 28.27 
        & 20.88 
        & 43.01 
        & 30.18 
        & \textbf{51.36} \\



        \midrule

        \multirow{6}{*}{Qwen-2.5}
        & 7B 
        & $\mathcal{M}_{T}$ 
        & 28.39 
        & 27.83 
        & 22.91 
        & 45.53 
        & 31.16 
        & 38.75 \\ 

        \cmidrule{2-9}

        & \multirow{7}{*}{3B}
        & $\mathcal{M}_{S}$ 
        & 16.02 
        & 14.10 
        & 21.75 
        & 37.98 
        & 22.46 
        & 36.29 \\

        & 
        & SFT (Non-Chat) 
        & 17.39 
        & 15.65 
        & 22.03 
        & 40.64 
        & 23.93 
        & 35.15 \\

        & 
        & SFT (Chat) 
        & 26.34 
        & 25.24 
        & 21.50 
        & 43.52 
        & 29.15 
        & 41.67 \\

        & 
        & KD (Non-Chat) 
        & 17.42 
        & 15.61 
        & 21.79 
        & 41.00 
        & 23.96 
        & 34.93 \\

        & 
        & KD (Chat) 
        & 24.81 
        & 26.35 
        & 20.79 
        & 43.86 
        & 28.95 
        & \textbf{40.99} \\



        \bottomrule
        \end{tabular}
    }
     \caption{\textbf{Utility Evaluation results.} Utility is measured using ROUGE-L (R-L) on DollyEval, SelfInst, VicunaEval, and S-NI, with AVG reporting the mean score across the four benchmarks. Safety is evaluated on SORRY-Bench, where lower values indicate safer behavior. The Method column indicates the training setting: $\mathcal{M}_{T}$ denotes the fine-tuned teacher model, $\mathcal{M}_{S}$ denotes the base instruct-tuned student model, SFT denotes supervised fine-tuning, and KD as standard distillation. }
    \label{tab:result_prompt_template_utility}
\end{table*}

%% file: assets/table/other_distillation_objective.tex
\begin{table}[!h]
\centering
\small

\vspace{3pt}
\resizebox{\columnwidth}{!}{%
\begin{tabular}{lccc}
\toprule
\textbf{Method} & \textbf{SORRY $\downarrow$} & \textbf{HarmBench $\downarrow$} & \textbf{Avg Utility $\uparrow$} \\
\midrule
\multicolumn{4}{l}{\textbf{LLaMA}} \\
RKL (Non-Chat)       & 28.11 & 12.56 & 24.79 \\
RKL (Chat)           & 30.31 & 26.63 & 29.98 \\
FKL+RKL (Non-Chat)   & 26.78    & 12.5 & 24.81 \\
FKL+RKL (Chat)       & 30.42    & 26.38 & 29.29 \\
\midrule
\multicolumn{4}{l}{\textbf{Gemma}} \\
RKL (Non-Chat)       & 18.86 & 13.75 & 23.83 \\
RKL (Chat)           & 41.29 & 16.88 & 30.63 \\
FKL+RKL (Non-Chat)   & 18.94    & 13.56 & 23.55 \\
FKL+RKL (Chat)       & 38.11    & 13.94 & 30.25 \\
\midrule
\multicolumn{4}{l}{\textbf{Qwen-2.5}} \\
RKL (Non-Chat)       & 32.96 & 18.38 & 23.75 \\
RKL (Chat)           & 40.00 & 29.38 & 29.03 \\
\bottomrule
\end{tabular}
}
\caption{\textbf{Generalization to other KD objectives under chat and non-chat templates.}
We report safety and utility for students trained with alternative distillation objectives under two prompt-template settings. Safety is measured using ASR (\%) on SORRY-Bench and HarmBench, where lower is better. Utility reports the average ROUGE-L score across the instruction-following benchmarks.}
\label{tab:other_distillation_objective}
\end{table}

%% file: assets/table/teacher_student_sft_template.tex
\begin{table*}[!t]
    \centering
    \small
    \vspace{5pt}
    \resizebox{0.8\linewidth}{!}{%
        \begin{tabular}{lrl|c|cc}
        \toprule
        \multirow{2}{*}{\textbf{Model}}
        & \multirow{2}{*}{\textbf{\#Params}}
        & \multirow{2}{*}{\textbf{Method}}
        & \multicolumn{1}{c|}{\textbf{Utility (R-L$\uparrow$)}}
        & \multicolumn{2}{c}{\textbf{Safety (ASR$\downarrow$)}} \\
        \cmidrule{4-6}
        & &
        & \textbf{AVG}
        & \textbf{SORRY}
        & \textbf{HarmBench} \\
        \midrule

        \multirow{6}{*}{LLaMA}
        & \multirow{3}{*}{8B}
        & $\mathcal{M}_{T}$
        & 20.20
        & 13.26
        & 8.69 \\
        \cmidrule{3-6}

        &
        & SFT (Non-Chat)
        & 24.86
        & 13.71$_{\textcolor{purple}{+0.45}}$
        & 8.63$_{\textcolor{teal}{-0.06}}$ \\

        &
        & SFT (Chat)
        & 30.33
        & 16.37$_{\textcolor{purple}{+3.11}}$
        & 15.69$_{\textcolor{purple}{+7.00}}$ \\

        \cmidrule{2-6}

        &
        \multirow{3}{*}{3B}
        & $\mathcal{M}_{S}$
        & 19.84
        & 26.72
        & 12.18 \\
        \cmidrule{3-6}

        &
        & SFT (Non-Chat)
        & 24.27
        & 27.12$_{\textcolor{purple}{+0.40}}$
        & 12.19$_{\textcolor{purple}{+0.01}}$ \\

        &
        & SFT (Chat)
        & 29.10
        & 28.71$_{\textcolor{purple}{+1.99}}$
        & 22.50$_{\textcolor{purple}{+10.32}}$ \\

        \midrule

        \multirow{6}{*}{Gemma}
        & \multirow{3}{*}{9B}
        & $\mathcal{M}_{T}$
        & 22.91
        & 12.05
        & 14.13 \\
        \cmidrule{3-6}

        &
        & SFT (Non-Chat)
        & 25.93
        & 12.58$_{\textcolor{purple}{+0.53}}$
        & 15.06$_{\textcolor{purple}{+0.93}}$ \\

        &
        & SFT (Chat)
        & 35.06
        & 46.36$_{\textcolor{purple}{+34.31}}$
        & 24.50$_{\textcolor{purple}{+10.37}}$ \\

        \cmidrule{2-6}

        &
        \multirow{3}{*}{2B}
        & $\mathcal{M}_{S}$
        & 22.35
        & 16.04
        & 11.56 \\
        \cmidrule{3-6}

        &
        & SFT (Non-Chat)
        & 23.66
        & 28.71$_{\textcolor{purple}{+12.67}}$
        & 20.37$_{\textcolor{purple}{+8.81}}$ \\

        &
        & SFT (Chat)
        & 30.05
        & 53.18$_{\textcolor{purple}{+37.14}}$
        & 13.25$_{\textcolor{purple}{+1.69}}$ \\

        \bottomrule
        \end{tabular}
    }
    \caption{\textbf{Chat-template SFT degrades safety across teacher and student models.} Utility is reported using AVG ROUGE-L across DollyEval, SelfInst, VicunaEval, and S-NI. Safety is evaluated using SORRY-Bench and HarmBench, where lower ASR indicates safer behavior. The subscripted deltas show the change relative to each model's own instruct-tuned baseline ($\mathcal{M}_{T}$ or $\mathcal{M}_{S}$).}
    \label{tab:abl_sft_teacher_student}
\end{table*}